%% file: main.tex
\documentclass{article}

\pdfoutput=1

\usepackage{PRIMEarxiv}

\usepackage[utf8]{inputenc} 
\usepackage[T1]{fontenc}    
\usepackage{hyperref}       
\usepackage{url}            
\usepackage{booktabs}       
\usepackage{amsfonts}       
\usepackage{nicefrac}       
\usepackage{microtype}      
\usepackage{lipsum}
\usepackage{fancyhdr}       
\usepackage{graphicx}       
\graphicspath{{media/}}     

\usepackage{amsmath}
\usepackage{multirow}
\usepackage{algorithm}
\usepackage{algpseudocode}
\usepackage{tikz}
\usepackage{tikz-qtree}
\usetikzlibrary{trees,decorations.text,math,calc, positioning, arrows.meta}
\usepackage[edges]{forest}
\usepackage{xcolor,stfloats}
\usepackage{color}

\definecolor{CY}{RGB}{233,246,254}
\definecolor{CornflowerBlue}{RGB}{100,149,237}
\definecolor{output-black}{RGB}{0,0,0}
\definecolor{output-white}{RGB}{255,255,255}
\tikzstyle{edge}=[-latex',draw=black!90,shorten <=1pt,shorten >=1pt]
\tikzstyle{redge}=[latex'-,draw=black!90,shorten <=1pt,shorten >=1pt]
\tikzstyle{dedge}=[latex'-latex',draw=black!90,shorten <=1pt,shorten >=1pt]

\tikzstyle{block}=[draw, text width=5em,align=center,shape=rectangle, rounded corners, , align=center]
\tikzstyle{nobox}=[align=center]

\title{Large Language Models Meet Graph Neural Networks: A Perspective of Graph Mining}

\author{
  Yuxin You \\
  School of Computer Science and Engineering \\
  University of Electronic Science and Technology of China\\
  Chengdu 611731, China\\
  \texttt{202321081209@std.uestc.edu.cn} \\
   \And
  Zhen Liu \\
  School of Computer Science and Engineering \\
  University of Electronic Science and Technology of China\\
  Chengdu 611731, China\\
  \texttt{quake@uestc.edu.cn} \\
   \And
  Xiangchao Wen \\
  School of Computer Science and Engineering \\
  University of Electronic Science and Technology of China\\
  Chengdu 611731, China\\
  \texttt{cha0s101cs@gmail.com} \\
   \And
  Yongtao Zhang \\
  School of Computer Science and Engineering \\
  University of Electronic Science and Technology of China\\
  Chengdu 611731, China\\
  \texttt{202222080730@std.uestc.edu.cn} \\
   \And
  Wei Ai \\
  54th Research Institute of CETC \\
  Shijiazhuang 050081, China\\
  \texttt{aiwei2001@sina.com} \\
}

\begin{document}
\maketitle

\begin{abstract}
Graph mining is an important area in data mining and machine learning that involves extracting valuable information from graph-structured data. In recent years, significant progress has been made in this field through the development of graph neural networks (GNNs). However, GNNs are still deficient in generalizing to diverse graph data. Aiming to this issue, Large Language Models (LLMs) could provide new solutions for graph mining tasks with their superior semantic understanding. In this review, we systematically review the combination and application techniques of LLMs and GNNs and present a novel taxonomy for research in this interdisciplinary field, which involves three main categories: GNN-driving-LLM, LLM-driving-GNN, and GNN-LLM-co-driving. Within this framework, we reveal the capabilities of LLMs in enhancing graph feature extraction as well as improving the effectiveness of downstream tasks such as node classification, link prediction, and community detection. Although LLMs have demonstrated their great potential in handling graph-structured data, their high computational requirements and complexity remain challenges. Future research needs to continue to explore how to efficiently fuse LLMs and GNNs to achieve more powerful graph learning and reasoning capabilities and provide new impetus for the development of graph mining techniques.
\end{abstract}


\section{Introduction}

Graphs are data structures used to describe relationships between objects, and they are widely used in many domains, such as social networks ~\cite{Nguyen_2020}, computer networks, and molecular structures. Some of these graphs contain hundreds of millions of node information, but the vast majority are redundant and irrelevant. What graph mining investigates is how to extract relevant or valuable knowledge and information from scaled graph data by using graph mining models.

Over the past decade, graph mining techniques have evolved, yielding many significant results and significantly contributing to the development of the field. Early research was inspired by the word vector technique ~\cite{mikolov2013efficientestimationwordrepresentations}, and Perozzi et al.~\cite{perozzi2014deepwalk} firstly proposed the random walk-based graph embedding method DeepWalk; Node2Vec~\cite{grover2016node2vecscalablefeaturelearning} proposed by Grover and Leskovec et al. generates node embeddings through biased random walk strategies for tasks such as node classification and link prediction, and Graph2Vec~\cite{narayanan2017graph2veclearningdistributedrepresentations}, on the other hand, embeds the entire graph into the vector space for graph classification tasks. For more efficient graph representation learning, the Graph Neural Networks (GNNs)~\cite{4700287}proposed by Kipf and Welling pioneered a new research paradigm. GNNs efficiently capture and collect structural information and dependencies in graph-structured data through information propagation and aggregation mechanisms, which enable the model to make precise predictions in complex graph structures. In recent years, various GNN architectures have been developed in information propagation and aggregation methods. For example, Graph Convolutional Network (GCN)~\cite{kipf2017semisupervisedclassificationgraphconvolutional} processes graph data through spectral convolution, which is widely used in node classification and graph classification tasks; Graph Attention Network (GAT)~\cite{veličković2018graphattentionnetworks} introduces an attention mechanism to adaptively assign the importance of different neighboring nodes, which enhances the expressive power of the model; GraphSAGE~\cite{hamilton2018inductiverepresentationlearninglarge} can efficiently process large-scale graph data by sampling and aggregating features from neighboring nodes. In addition, for heterogeneous graph analysis, the Heterogeneous Graph Attention Network (HAN)~\cite{wang2021heterogeneousgraphattentionnetwork} proposed by Wang et al. aggregates information and learns between different types of nodes and edges through the attention mechanism, while Zhang et al.'s Heterogeneous Graph Neural Network (HetGNN)~\cite{10.1145/3292500.3330961} learns embeddings by sampling various types of nodes and edges, which enables it to manage complex heterogeneous graph structures. These results not only demonstrate the prospect of wide application of graph mining techniques but also promote the development of related research fields.

\begin{figure}[tbp]
\centering
\includegraphics[width=9 cm]{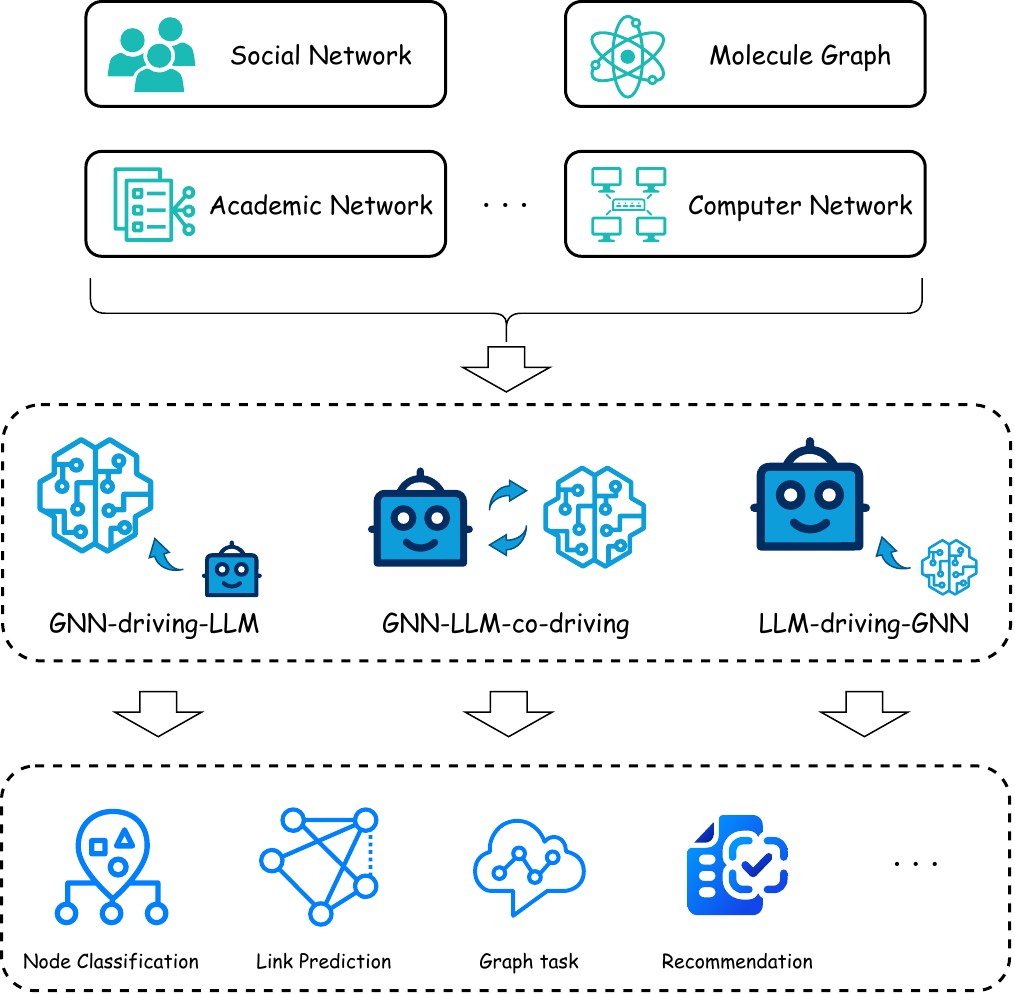}
\caption{Based on the relationship between LLMs and GNNs, we categorize the application scenarios of LLMs in graph mining into three groups: GNN-driving-LLM, LLM-driving-GNN, and GNN-LLM-co-driving. These models are capable of handling a variety of datasets and achieving success in various downstream tasks.}
\label{fig:motives}
\end{figure}
\unskip

In 2022, the emergence of Large Language Models (LLMs), represented by ChatGPT ~\cite{10.1162/daed_a_01905}, revolutionized the field of Natural Language Processing (NLP) as well as the domain of artificial intelligence research. The large language model is pre-trained on a large amount of text, which can learn rich language expressions and huge amount of real-world knowledge, and has excellent semantic understanding capabilities~\cite{petroni2019languagemodelsknowledgebases}. For example, BERT~\cite{devlin2019bertpretrainingdeepbidirectional}  uses a bidirectional attention mechanism to capture the contextual information of the text, which enables the model to perform on various NLP tasks such as question answering, named entity recognition, and sentence classification. GPT-3~\cite{brown2020languagemodelsfewshotlearners}, with 175 billion parameters, is pre-trained by an autoregressive approach and is capable of performing a wide range of tasks from text generation and translation to dialog systems. Its strength lies in its ability to perform new tasks without fine-tuning and with a small number of examples or hints, demonstrating impressive zero-shot and few-shot learning capabilities. These LLMs can be applied to various downstream tasks with little additional training. Although large language models were initially developed for natural language processing, researchers have also been exploring the application of LLMs on multimodal data in recent years. For instance, the DALL-E~\cite{ramesh2021zeroshottexttoimagegeneration} model that combines image and text can generate corresponding images based on the text by performing self-supervised learning on paired image-text corpora. These multimodal models demonstrate outstanding potential for processing and understanding different data types.

Given the disruptive potential of large models, many researchers in graph mining have focused on them in the last two years, expecting them to bring new developments to the field of graph mining. Zhang et al.~\cite{zhang2023graphmeetsllmslarge} discusses the challenges and opportunities presented by the combination of graphs and LLMs and showcases the potential of these models across various application domains. Chen et al.~\cite{chen2024exploringpotentiallargelanguage} explores the potential of utilizing LLMs in graph learning tasks and investigates two possible approaches: LLMs-as-Enhancers and LLMs-as-Predictors. Liu et al.~\cite{liu2024graphfoundationmodelssurvey} introduced the concept of Graph Foundation Models (GFMs) and provided a taxonomy and review of existing work related to GFMs.These studies suggest that integrating LLMs with graph neural networks can enhance various downstream graph mining tasks. The reason is that, although GNNs excel at capturing structural information, they have limitations in terms of expressiveness and generalizability\cite{yang2023individualstructuralgraphinformation}, and their semantically constrained embeddings often fail to characterize the node features for complex node information fully. On the contrary, LLMs are good at processing complex texts but are often deficient in structural information processing. Combining the strengths of both can significantly improve the accuracy of graph mining.

To the best of our knowledge, existing reviews on the application of large models in graph mining mainly focus on categorizing LLMs simply as enhancers or predictors~\cite{chen2024exploringpotentiallargelanguage}. However, such a categorization approach is only a simple and superficial combination of LLMs and GNNs considered as independent models. It fails to delve into the profound fusion potential of LLMs and GNNs in the graph mining domain. Therefore, we propose a new classification framework, as shown in Figure \ref{fig:motives}, based on the main driving components in graph mining models, which are categorized into three groups: GNN-driving-LLMs, LLM-driving-GNNs, and GNN-LLM-co-driving. In this new classification framework, the GNN-driving-LLM mode emphasizes the GNN as the central task processing module, and the LLM plays an assisting role in specific tasks or scenarios, such as natural language interpretation or feature extraction. On the contrary, the LLM-driving-GNN mode places the LLM at the core and the GNN as an auxiliary tool for processing and guiding the graph-structured data to enhance the model's performance in complex graph data. In the GNN-LLM-co-driving mode, on the other hand, GNN and LLM work closely together to form a kind of interdependent joint model that collaboratively solves the corresponding graph mining tasks. Such a classification not only helps to fully understand the deep integration of LLMs and GNNs but also provides new ideas for future research directions.

\section{preliminary}

\subsection{Graph Mining}

Graph mining tasks are important in data knowledge discovery, aiming to extract valuable information from graph-structured data. Graph-structured data consists of nodes and edges, where nodes represent entities and edges represent relationships between represented entities. It is widely applied in fields such as social network analysis ~\cite{2014Social}, recommender systems ~\cite{2015Recommendation}, chemistry ~\cite{yan2002gspan} and knowledge graphs ~\cite {2021A}.

Common graph mining tasks include node classification, link prediction, graph clustering, graph matching, community detection, frequent subgraph mining, etc. Node classification utilizes labeled nodes as training data and predicts the classes of unlabeled nodes in the graph, such as node embeddings based on random walks ~\cite{grover2016node2vecscalablefeaturelearning}, Graph Convolutional Networks (GCN)~\cite{ kipf2017semisupervisedclassificationgraphconvolutional}. Link prediction is used to predict potential future edges, with important applications in recommender systems (e.g., friend recommendation, item recommendation). Common link prediction algorithms include neighborhood-based algorithms ~\cite{ liben2003link}, path-dependent models ~\cite{backstrom2011supervised}, deep learning methods ~\cite{zhang2018link}, and so on. Graph clustering ~\cite{schaeffer2007graph} and graph matching ~\cite{conte2004thirty} are respectively concerned with the grouping of nodes and subgraph correspondence between different graphs. Graph clustering groups graph nodes such that nodes within the same group are more tightly connected, while graph matching finds corresponding subgraphs between different graphs, which is commonly used in chemical molecular structure comparison and pattern recognition. Community detection ~\cite{fortunato2010community} identifies subsets or associations in the graph, where nodes within a community are closely connected but have relatively few connections to nodes outside the community; commonly used algorithms such as the Girvan-Newman algorithm ~\cite{girvan2002community}, Louvain's method ~ \cite{blondel2008fast} and so on. Frequent subgraph mining ~\cite{inokuchi2000apriori} spots frequently occurring subgraphs from a set of graphs, which can be used in chemoinformatics to discover common molecular structures. By analyzing a large number of molecular diagrams, frequent subgraph mining can reveal which molecular fragments appear repeatedly across multiple compounds, providing valuable insights for fields such as bioinformatics and chemoinformatics~\cite{borgelt2002mining}.

\subsection{Large Language Model}

A large language model consists of a neural network with many parameters (usually billions of weights or more). In recent years, thanks to the introduction of the Transformer architecture and its ability to be pre-trained on large-scale text data, it has made significant progress in natural language processing.

The transformer model ~\cite{vaswani2023attentionneed} proposed by Vaswani et al. in 2017 is based on the attention mechanism, which can greatly improve training efficiency and performance. Its fundamental unit is the multi-head self-attention mechanism, which can be expressed by the following formula:
\begin{equation}
\text{Attention}(Q, K, V) = \text{softmax}\left(\frac{QK^T}{\sqrt{d_k}}\right)V, 
\end{equation}
Where \(Q\), \(K\), and \(V\) denote the Query, Key, and Value matrices, respectively, and \(d_k\) is the dimension of the key vector. The multiple head mechanism further computes several different attention values and combines them:
\begin{equation}
\text{MultiHead}(Q, K, V) = \text{Concat}(\text{head}_1, \text{head}_2, \ldots, \text{head}_h)W^O.
\end{equation}
The types and applications of LLMs are constantly evolving with the development of architectures and training methods. Current LLMs can be classified into the categories of autoregressive models, masked language models, encoder-decoder models, contrastive learning models, multimodal models, and others. Autoregressive models generate text by predicting the next word in a sequence, and such models can generate coherent text and perform very well in generative tasks such as GPT~\cite{radford2018improving}, GPT-2~\cite{radford2019language}, GPT-3~\cite {brown2020languagemodelsfewshotlearners} and GPT-4~\cite{openai2024gpt4technicalreport}.On the other hand, Masked language models typically mask the words in a given utterance to train the model to predict these masked words and thus learn the contextual representation. Representative models of masked language models include BERT~\cite{devlin2019bertpretrainingdeepbidirectional}, RoBERTa~\cite{liu2019roberta} and XLNet~\cite{yang2020xlnetgeneralizedautoregressivepretraining}. The encoder-decoder models achieve a high degree of parallelization and dramatically improve computational efficiency by encoding the input text into a contextual representation and decoding it to generate the target text. Classic models of this type include BART~\cite{lewis2019bartdenoisingsequencetosequencepretraining}, and T5~ \cite{raffel2020exploring}.Contrastive learning models, such as SimCLR~\cite{chen2020simpleframeworkcontrastivelearning}, train the model to differentiate between positive and negative samples by constructing pairs of positive and negative samples so that the model can capture relevant features and similarities in the data. Finally, multimodal models are able to process information from multiple modalities (including images, videos, and texts) to enhance the performance of the model in multimodal tasks, such as CLIP~\cite{radford2021learningtransferablevisualmodels} and DALL-E~\cite{ ramesh2021zeroshottexttoimagegeneration}. The continuous development of large language models not only enhances the ability of computers in natural language processing but also provides new ideas and methods for solving broader and more complex data analysis and processing tasks.

\section{Techniques of the LLMs combined with GNNs}
Regarding techniques of the LLMs combined with GNNs, as shown in Figure \ref{fig:tree}, we propose a novel taxonomy including GNN-driving-LLM, LLM-driving-GNN, and GNN-LLM-co-driving.

\input{fig/tree}

\subsection{GNN-driving-LLM}

In graph mining, GNNs are a class of deep learning models specialized in processing graph data. By combining the structural information and node features of the graph, they are able to effectively learn the representation of nodes, edges, and the overall graph in a graph. Text-attributed graphs (TAGs), in which the attributes of nodes exist in the form of text, are ubiquitous in the research of graph machine learning, such as product networks~\cite{mcauley2013hidden}, social networks~\cite{Nguyen_2020, granovetter1973strength} and citation networks~\cite{yao2019graph, price1965networks}, where textual attributes provide key semantic information for the graph mining task. Therefore, when dealing with such graphs, the structure of the graph, the textual information, and their interrelationships must be considered simultaneously.

Traditionally, the processing of node text attributes often relies on shallow embedding methods, such as the classic Cora dataset ~\cite{yang2016revisitingsemisupervisedlearninggraph}, which only provides embedding features based on bag-of-words models. This approach is coarse in semantic understanding, leading to the limited performance of GNNs in processing textual attribute graphs. However, with the development of LLMs, their powerful text processing and semantic understanding capabilities provide a new solution to this problem. LLMs can extract richer semantic features from the textual attributes of nodes and generate additional auxiliary information, such as attribute interpretations and pseudo-labels, which provide more resounding semantic support for node embedding in graph mining tasks. In previous research ~\cite{li2021adsgnnbehaviorgraphaugmentedrelevance, Zhu_2021}, techniques combining language model (LM) and GNNs have been investigated to use LMs for encoding and providing them as node features to GNNs, and the introduction of extensive language modeling further enhances the effectiveness of this approach.
\begin{figure}[tbp]
\centering
\includegraphics[width=15 cm]{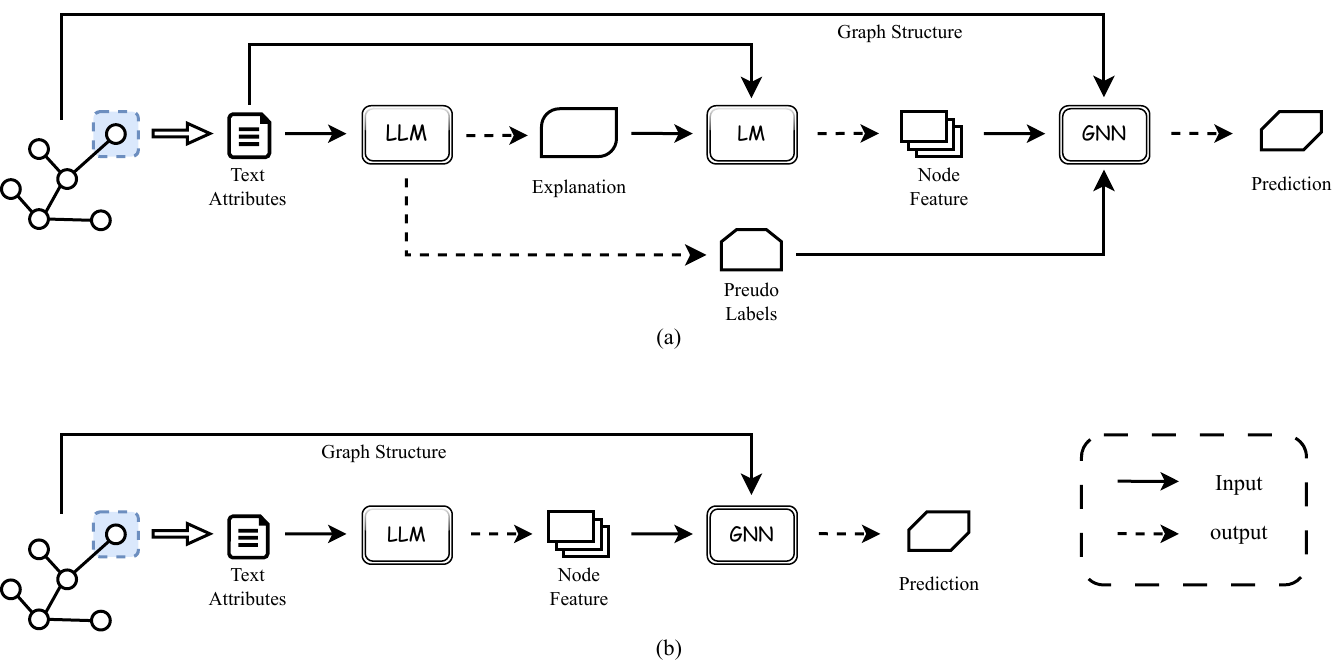}
\caption{GNN-driving-LLM: \textbf{a}) LLMs generate additional information for the textual attributes of the nodes. These features extracted by LLMs are fed into the next layer of the language model to generate enhanced node embeddings, which are finally processed by GNNs for downstream tasks; \textbf{b}) the textual embeddings exported by LLMs can be directly used as the initial node embeddings for GNNs.}
\label{fig:enhance}
\end{figure}
\unskip
In the GNN-driving-LLM model, GNNs are still used as the central information processing unit, but the text attributes are deeply parsed and embedded by incorporating LLMs, which makes the GNNs more accurate in capturing the semantic information of the nodes and thus improves the performance of the model in the downstream tasks. As shown in Figure \ref{fig:enhance}(a), LLMs generate additional information for the textual attributes of the nodes. Next, these features extracted by LLMs are fed into the next layer of the language model to generate enhanced node embeddings. A typical example is the TAPE~\cite{he2024harnessingexplanationsllmtolminterpreter} model, which first predicts the node text (such as paper titles and abstracts) in LLMs and generates the corresponding interpretations. Through task-specific prompts, LLMs generate categorization predictions and explanation texts. The interpreted text generated by LLMs is then fed into a smaller language model, processed through fine-tuning and transformed into node features (including original text features\(\mathbf{h}_{orig}\), explanation features\(\mathbf{h}_{expl}\) and prediction features\(\mathbf{h}_{pred}\)). 
The combination of these three classes of features as \(\mathbf{h}_{TAPE}\) is used for downstream GNNs training. Finally, the GNN models are trained on the generated features to achieve the node classification task. The following formula can describe the process:
\begin{equation}
\mathbf{s} = \text{LLM}(x, p), \quad
\mathbf{h}= \text{LM}(\mathbf{s} ,p) \in \mathbb{R}^{N \times d}, \quad
\hat{\mathbf{y}} =\text{GNN}(\mathbf{h}, \mathbf{s} , A) \in \mathbb{R}^{N \times C},
\end{equation}
where the LLMs accept a raw text \( x = (x_1, x_2, \ldots, x_q) \) as input and generates a sequence of interpreted text \( s = (s_1, s_2, \ldots, s_m) \) as output, with \( p \) is the cue for the LLMs. \( \mathbf{h} \in \mathbb{R}^{N \times d}\) is the output of LMs, which is the node embedding matrix augmented by LLMs, and \(A\in \mathbb{R}^{N \times N}\) is the adjacency matrix of the graph.
Several studies have shown that this strategy of combining LLMs and GNNs has significant advantages in practical applications. For example, LLMRec~\cite{wei2024llmreclargelanguagemodels} is the first work on graph enhancement using LLMs by augmenting user-item interaction edges, item node attributes, and user node profiles, which effectively solves the problems of data sparsity and low-quality auxiliary information in recommender systems. Similarly, RLMRec~\cite{Ren_2024} also utilizes LLMs to enhance representation learning in existing recommender systems, which uses LLMs to respectively process textual information of items, as well as user interaction information and textual attributes of related items, to generate item profiles and user profiles. By maximizing mutual information, RLMRec aligns the semantic representations from LLMs with collaborative relationship representations, significantly improving the performance of the recommendation system. PRODIGY~\cite{huang2023prodigyenablingincontextlearning} is a framework for pre-training context learners on prompt graphs. It leverages the powerful zero-sample capability of LLMs to encode textual information of graph nodes, enabling context learning on graphs. Inspired by the remarkable effectiveness of prompt learning in NLP, ALL-in-one~\cite{sun2023onemultitaskpromptinggraph} proposes a new multitask prompting approach that unifies graph prompts and language prompts formats by prompt tokens, token structures, and insertion patterns. It enables the NLP prompting concepts to be seamlessly introduced into the graph domain, thus unifying the formats of graph prompting and language prompting.

When using embeddable or open-source LLMs, the text embeddings they generate can be accessed directly. In this case, the LLMs first extract textual features for each node, which are then fed into the GNNs as initial node embeddings. Subsequently, the GNNs combine the graph structure information with these augmented node embeddings through their message passing and feature aggregation mechanisms, as shown in Figure \ref{fig:enhance}(b). In general, the process can be described by the following equation:
\begin{equation}
\mathbf{h}= \text{LLM}(\mathbf{s}) , \quad
\hat{\mathbf{y}} =\text{GNN}(\mathbf{h}, A) .
\end{equation}
This approach significantly improves the performance of graph mining tasks, especially in downstream tasks such as node classification, link prediction, and graph classification. OFA~\cite{liu2024alltraininggraphmodel} embeds textual descriptions of graph datasets from different domains into the same feature space, thus becoming the first cross-domain generalized graph classification model. GaLM~\cite{xie2023graphawarelanguagemodelpretraining} pretrains on large-scale heterogeneous graphs and incorporates structural information into the fine-tuning stage of LLM to generate higher quality node embeddings, thus improving performance for downstream applications in different graph patterns. LEADING~\cite{ xue2023efficientlargelanguagemodels} is an end-to-end fine-tuning algorithm that significantly improves LLMs' computational and data efficiency in processing TAGs by reducing coding redundancy and propagation redundancy. GraphEdit~\cite{guo2024grapheditlargelanguagemodels} enhances LLMs' reasoning ability on node relationships in graph data through instruction tuning to solve the noise and sparsity problems in graph structures. Specifically, GraphEdit first uses LLMs to generate semantic embeddings of nodes and filter candidate edges by a lightweight edge predictor; then, in combination with the original graph structure, it uses the inference ability of LLMs to optimize edge additions and deletions and generates the improved graph structure. Eventually, the optimized graph structure is used for GNNs training to support downstream tasks such as node classification, thus realizing denoising and global dependency mining of the graph structure.

With their exceptional ability to process text sequences, LLMs perform excellently in handling TAGs. They can extract deep semantic information from textual attributes, providing richer feature representations than traditional methods. In addition, rather than traditional GNNs that need to design different architectures for different datasets, combining LLMs and GNNs can process data from different domains by unifying the feature space and cross-domain embedding methods. This approach, which combines the diversity of language models and the ability to understand the structure of graph neural networks, offers flexibility in dealing with complex attributes and structures.

Beyond that, LLMs can also improve the performance of downstream tasks through data augmentation.  For example, LLM-GNN\cite{chen2024labelfreenodeclassificationgraphs} achieves efficient and low-cost  label-free node classification through LLM-based zero-shot labeling and GNN-based extended learning. LLM4NG~\cite{ yu2024leveraginglargelanguagemodels} represents a typical application of graph generation learning. It utilizes a large language model to generate new nodes and integrates these nodes into the original graph structure via an edge predictor to generate a new graph structure. This approach significantly improves model performance in few-shot scenarios, demonstrating the effectiveness of augmenting model learning capabilities by generating samples. Similarly, OpenGraph~\cite{xia2024opengraphopengraphfoundation} uses LLMs for generating synthetic graph data (e.g., nodes and edges) as well as augmenting the pre-training data. It also employs a unified graph tokenizer and an efficient graph Transformer, achieving excellent generalization on multi-domain graph data in zero-shot learning.

Despite the fact that LLMs can enhance model performance, they are extremely demanding of computational resources, especially when processing large amounts of textual data, requiring significant computational resources or frequent API query calls to the LLMs, which comes with a high cost. Moreover, although LLMs provide rich semantic information, their decision-making process is often opaque and lacks interpretability. Therefore, how to efficiently fuse text embedding with graph structure information in the framework of GNNs remains a challenging issue that requires further research.

\subsection{LLM-driving-GNN}

\begin{figure}[tbp]
\centering
\includegraphics[width=15 cm]{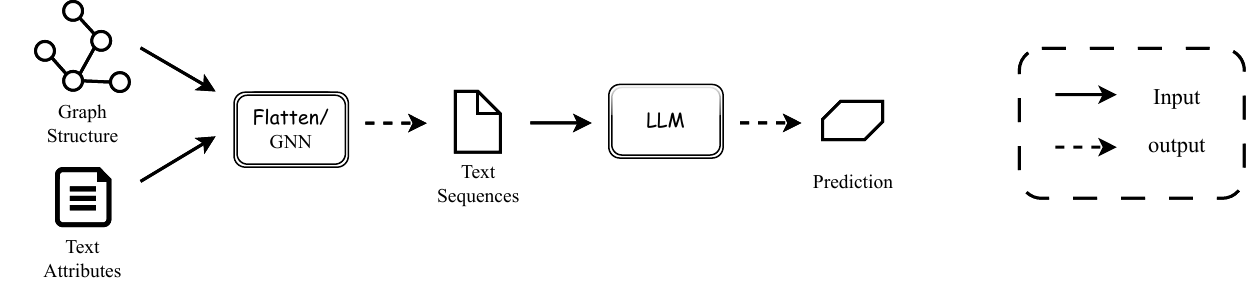}
\caption{LLM-driving-GNN: Models use spreading functions or GNNs to transform graph data into sequence text so that LLMs can directly understand the graph data and accomplish downstream tasks.  }
\label{fig:llm}
\end{figure}
\unskip

In some graph mining tasks, LLMs, with their powerful zero-shot learning capabilities, can directly perform prediction, classification, or reasoning. The computational power of LLMs and the advantages of deep learning algorithms enable them to perform well in these tasks. However, since LLMs can only accept sequential text as input, an additional step is required for graph data: transforming the graph data, which has been defined in various ways in terms of structure and features, into sequential text directly fed into LLMs.
In recent years, many studies have explored the applicability of LLMs for tasks downstream of graph structures, yielding some preliminary results~\cite{ liu2023chatgptgoodrecommenderpreliminary, creswell2022selectioninferenceexploitinglargelanguage, ji2023exploringchatgptsabilityrank}. These results suggest that LLMs have achieved initial success in processing implicit graph structures. Wang et al.\cite{wang2024languagemodelssolvegraph} proposed a synthetic benchmark, NLGraph, for evaluating the performance of LLMs in the task of reasoning about graph structures. It was found that the capability of LLMs has been demonstrated in simple graph reasoning tasks but is still insufficient when dealing with complex problems.Additionally, several studies have explored the capability of LLMs in processing graph-structured data and their effectiveness in extracting and utilizing graph structural information~\cite{huang2024llmseffectivelyleveragegraph, hu2023textdeepdivelarge}. These studies have opened new avenues for exploiting the capabilities of LLMs in graph applications and further exploring the incorporation of structured data into LLMs frameworks. A growing number of researchers are committed to exploring how to better apply LLMs to downstream tasks containing graph structures and further improve their performance.

Typically, to enable LLMs to understand graph data, researchers use specific methods to convert graph data into textual descriptions, which are then used as inputs to LLMs and finally extract predictions from the output of the model. Among these methods, flattening functions can directly convert graphs into textual descriptions, which is more convenient and intuitive, while GNNs have demonstrated excellent ability in understanding graph structures through information propagation and aggregation among nodes. Therefore, researchers have also explored using GNNs to transform graph data with different structures and features into sequential text to fully leverage the structural information in the graph data for prediction. As shown in Figure \ref{fig:llm}, the method of encoding graph structure data into text for use in LLMs was first comprehensively investigated by Fatemi et al.~\cite{fatemi2023talklikegraphencoding}, and can be described by the following equation:
\begin{equation}
    \text{Describe} = f(G, Attr), \quad
    \mathbf{A} = \text{LLM}(\text{Describe}, p),
\end{equation}
where \(f\) represents the flattening function or GNNs, which inputs the graph \(G\) and the text attributes \(Attr\) on each node or edge of the graph to get the text description, i.e., \(\text{Describe}\), to be fed to the LLMs.  \(p\) denotes the prompts, and \(\mathbf{A}\) is the answers given by the LLMs, from which the predicted labels of downstream tasks can be extracted in a specific way. Research has shown that choosing a suitable graph encoding method can significantly improve the performance of LLMs in graph reasoning tasks. Thus, one of the current research priorities is to explore suitable graph encoding methods. For example, GPT4Graph~\cite{guo2023gpt4graphlargelanguagemodels} converts graph data into a Graph Description Language (GDL) like GraphML~\cite{brandes2013graph}. It generates prompts in conjunction with user queries so that the large language model can understand and process the graph-structured data. GraphText~\cite{zhao2023graphtextgraphreasoningtext} constructs a graph syntax tree from graph data and generates graph prompts through traversal of the tree, expressing them in natural language so that LLMs can treat graph reasoning as a text generation task. An alternative approach is to record the graphical data directly in natural language, i.e., to describe the graphical data with a digitally organized list of nodes and edges~\cite{liu2023evaluatinglargelanguagemodels}. GraphGPT~\cite{tang2024graphgptgraphinstructiontuning} also utilizes Chain-of-Thought (CoT) ~\cite{wei2022chain} to augment the model's reasoning capabilities. However, in terms of transforming graph data, GraphGPT trains a lightweight graph-text projector that is able to align representations between text and graph structure, allowing the model to switch seamlessly during processing. Similarly, MoleculeSTM~\cite{cao2023instructmolmultimodalintegrationbuilding} uses a graph encoder as a molecular graph structure encoder. The method enables large language models to align molecular structures to natural language by aligning molecular graphs and textual representations through contrastive training and then employing a lightweight alignment projector to map graph features into the word embedding space. DGTL~\cite{qin2024disentangledrepresentationlearninglarge} encodes the raw textual information in TAGs using a frozen LLM and then captures the graph neighborhood information in TAGs by a set of custom disentangled graph neural network layers. Finally, the features learned from these disentangled layers are used to fine-tune the LLMs to help the model better understand the complex graph structure information in the TAGs, thereby improving the final prediction ability of the LLMs. On the other hand, GraphTranslator~\cite{zhang2024graphtranslatoraligninggraphmodel} proposes a mechanism called Producer for creating graph-text alignment data, which enables a large language model to predict graph data based on language instructions. HiGPT~\cite{tang2024higptheterogeneousgraphlanguage} adapts to diverse heterogeneous graph learning tasks without downstream fine-tuning through a heterogeneous graph instruction-tuning paradigm.

Another way to enhance the capability of LLMs on graph-structured data is by fine-tuning LLMs to enhance their graph mining capabilities. Several researchers have explored this area and made notable progress. GIMLET~\cite{zhao2023gimletunifiedgraphtextmodel} fine-tunes LLMs to output predictive labels directly, thus providing accurate predictions without additional parsing steps. MuseGraph~\cite{tan2024musegraphgraphorientedinstructiontuning} generates compact graph descriptions using neighbor nodes and random walks and creates task-based CoT instruction sets to fine-tune the large language model. The method dynamically allocates instruction packages between tasks and datasets to ensure the effectiveness and generalization of the training process. Eventually, the graph structure data is converted into a format suitable for LLMs, which allows the fine-tuned model to fit downstream tasks such as node classification, link prediction, and graph-to-text generation. InstructGLM~\cite{ye2024languagegraphneeds} designed a series of rule-based, highly scalable natural language prompts for describing graph structures and performing graph tasks, and fine-tunes large language models with these instructions, enabling them to understand and process these descriptions to perform graph tasks. GraphLLM~\cite{chai2023graphllmboostinggraphreasoning} adopts an end-to-end approach that integrates a graph learning module (graph transformer) with LLMs. Specifically, the approach uses a textual Transformer encoder-decoder to extract the necessary information from the node descriptions, learns the graph structure through the graph Transformer, and generates overall graph representations by aggregating the node representations. Ultimately, these graph representations are used to generate graph-enhanced prefixes injected in each LLM attention layer. This approach allows the LLM to work synergistically with the graph Transformer to incorporate structural information critical for graph inference and thus improve performance on graph reasoning tasks.

Unlike the above approaches, the Graph-ToolFormer framework ~\cite{zhang2023graphtoolformerempowerllmsgraph} uses API calls to invoke external graph inference tools to complete reasoning tasks. First, ChatGPT is utilized to annotate and expand the manually written graph reasoning task prompts to generate a large dataset of prompts containing graph reasoning API calls. Then, the generated dataset is used to fine-tune pre-trained causal LLMs (e.g., GPT-J~\cite{gpt-j} and LLaMA~\cite{touvron2023llamaopenefficientfoundation}) and teach them how to use external graph inference tools in the generated output. Finally, the fine-tuned Graph-ToolFormer models are able to automatically add the corresponding graph reasoning API calls to the output statements when they receive input queries and questions. Through the above steps, the Graph-ToolFormer framework realizes the ability to empower existing LLMs to handle complex graph reasoning tasks, effectively addressing the current limitations of LLMs in handling precise computation, multi-step logical reasoning, spatial and topological awareness, etc. LLMs can also be used to generate new GNN architectures. Wang et al.~\cite{wang2024graphneuralarchitecturesearch} proposed a novel graph neural network architecture search method, GPT4GNAS, which guides GPT-4 to understand the search space and search strategies of GNAS by designing a new type of prompt. These prompts are iteratively run to generate new GNN architectures, and the evaluation results are used as feedback to optimize the generated architectures further. ChatRule~\cite{luo2024chatrulemininglogicalrules} utilizes LLMs to mine logical rules as well as learn and represent graph structures by combining semantic and structural information from Knowledge Graphs (KGs), helping encode and process graph structures. These rules can be regarded as new graph structures to construct new knowledge graphs by capturing the generative rules and patterns of graphs. 
Meanwhile, GNP~\cite{tian2023graphneuralpromptinglarge}  extracts valuable knowledge from knowledge graphs, and through graph neural network encoding, cross-modal pooling, and self-supervised learning, it significantly improves the performance of LLMs in common-sense reasoning and biomedical reasoning tasks. 

In conclusion, when LLMs dominate downstream tasks such as prediction, classification, and reasoning, they demonstrate significant advantages over traditional GNNs, especially in zero-sample learning and processing textual attributes. LLMs can utilize their powerful text generation and comprehension capabilities to predict and classify the graph data directly without the complex structural processing required by GNNs. However, since every coin has two sides, each approach requires careful trade-offs of the advantages and disadvantages between LLMs and GNNs. Converting graph data into textual descriptions can simplify the processing flow and enable LLMs to utilize their textual processing capabilities for reasoning directly. Nevertheless, due to the input length limitation, this approach may result in the loss of graph structure information, and the text conversion process is so complex that it is unsuitable for processing complex graph data. On the other hand, Combining GNNs can fully utilize the information in the graph structure and enhance the processing capability of the model. However, this integration method increases the complexity of the system. Effective integration of GNNs with LLMs requires careful design and tuning to ensure that both can effectively work together to handle complex graph data. To sum up, how to efficiently combine graph structural information with LLMs for more powerful graph learning and reasoning is the core of research in this area.

\subsection{GNN-LLM-co-driving}

\begin{figure}[tbp]
\centering
\includegraphics[width=15 cm]{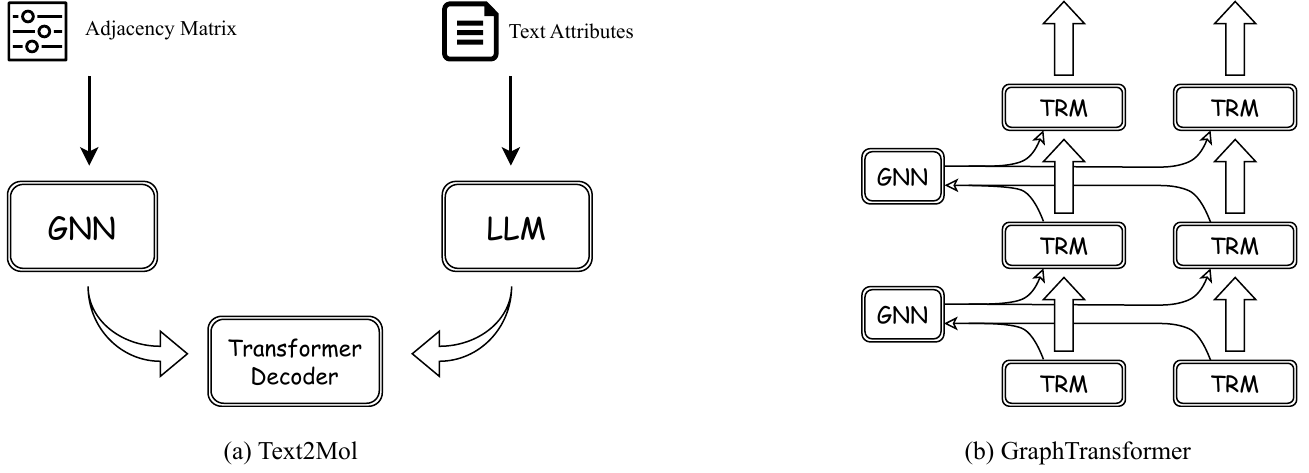}
\caption{\textbf{a}) Text encoding and graph aggregation are iteratively executed through hierarchical Graph Neural Networks (GNN) and Transformers (TRM); \textbf{b}) Data from these two different modalities are projected into an aligned semantic embedding space, where attention mechanisms are used to learn the correlation rules between molecular substructures and textual keywords.}
\label{fig:co_driving}
\end{figure}
\unskip

In GNN-driving-LLM, LLMs act as the preprocessors of graph data, primarily converting the textual attributes of the graph into rich feature representations through their powerful text comprehension and generation capabilities and then passing these representations to GNNs for further structural processing. In this setup, LLMs play a role in information extraction and feature enhancement. In contrast, in LLM-driving-GNN, GNNs are mainly responsible for processing graph structural information and utilizing this structural information to enhance the prediction, classification, reasoning, and generation capabilities of LLMs. 

GNN-LLM-co-driving synthesizes the strengths of GNNs and LLMs, leveraging their collaboration to solve complex tasks. Different from GNN-driving-LLM and LLM-driving-GNN, the co-driving strategy emphasizes the deep interaction and complementarity between GNNs and LLMs. In this architecture, GNNs and LLMs co-drive the learning process of the model, alternating and complementing each other. GNNs, with their advantage in graph structural information,  help LLMs generate semantically deeper features under complex structures; at the same time, LLMs provide GNNs with more accurate node and edge representations through their strong ability in text sequence processing. This bidirectional interaction not only improves the model's comprehensive processing capability on graph structures and textual information but also enhances the robustness and generalization of the overall model, which can better cope with diverse tasks and datasets.

Typical co-driving models are the GraphFormers framework proposed by Yang et al. ~\cite{yang2023graphformersgnnnestedtransformersrepresentation}, as shown in figure \ref{fig:co_driving} a). It creates a unified architecture capable of processing textual and graph structural information simultaneously by embedding GNNs into each transformer layer. The graph is first aggregated in each layer through GNN, bringing neighborhood information to the central node. Then, the enhanced node features are processed by the transformer to generate richer node representations. This allows the text encoding and graph aggregation processes to alternate in the same workflow, allowing nodes to exchange information at each layer to enhance the node representations with the information from neighboring nodes. PATTON~\cite{jin2023pattonlanguagemodelpretraining} adopts the GraphFormers framework and designs two pretraining strategies—Network Contextualized Masked Language Modeling (NMLM) and Masked Node Prediction (MNP)—based on it. By jointly optimizing the two objective functions NMLM and MNP to carry out the pretraining process, this approach enhances the model’s semantic understanding at both the vocabulary and document levels, enabling the pre-trained language model to sufficiently understand and represent the complex semantic information in rich textual networks. Zhao et al.~\cite{zhao2023learninglargescaletextattributedgraphs} propose the GLEM model, which combines GNNs and LLMs and alternately updates LLMs and GNNs to generate pseudo-labels with each other through a Variational Expectation-Maximization (EM) framework. By integrating textual semantics and graph structural information, GLEM can effectively perform node representation learning on large-scale text-attributed graphs and improve node classification performance. GREASELM~\cite{zhang2022greaselmgraphreasoningenhanced} uses a modality interaction mechanism to facilitate bidirectional information transfer between each layer of the LM and the GNN, deeply integrating language context with knowledge graph representations to realize joint reasoning to answer complex questions. 

Unlike the aforementioned architectures that combine GNN and LLM models, Text2Mol~\cite{edwards-etal-2021-text2mol} employs two independent encoders for textual and molecular representations, respectively. By projecting the data from these two modalities into an aligned semantic embedding space, the model achieves a cross-modal search for retrieving molecules from natural language descriptions. To enhance the interpretability of the model, the Text2Mol framework introduces a cross-modal attention model based on the Transformer decoder. The model utilizes the output of SciBERT~\cite{beltagy2019scibertpretrainedlanguagemodel} as the source sequence and the node representations generated by the GCN model as the target sequence, learning the association rules between molecule substructures and text keywords through an attention mechanism. The architecture is shown in figure \ref{fig:co_driving} b). Similarly, methods such as MoleculeSTM~\cite{liu2024multimodalmoleculestructuretextmodel}, CLAMP~\cite{seidl2023enhancingactivitypredictionmodels}, ConGraT ~\cite{brannon2024congratselfsupervisedcontrastivepretraining}, G2P2~\cite{wen2024prompttuninggraphaugmentedlowresource}, GRENADE~\cite{li2023grenadegraphcentriclanguagemodel}  also employ independent GNN encoders and LLMs encoders to process molecular and textual data separately, and then map the embedded representations to a shared joint representation space for contrastive learning. However, these models differ in implementation details or in the specific application scenarios they are applicable to. MoleculeSTM focuses on solving new challenges in drug design, such as structure-text retrieval and text-based molecular editing, and constructs PubChemSTM, a multimodal dataset containing a large number of chemical structure-text pairs. CLAMP, on the other hand, is primarily dedicated to zero-sample bio-activity prediction. Pretraining on large-scale chemical databases (such as PubChem) containing molecular structures, text descriptions, and bioactivity measurement data significantly enhances the generalization ability of activity prediction models. ConGraT connects an adapter module after each of the text encoder and graph node encoders, respectively, which consists of two fully connected layers to generate text embeddings and graph node embeddings of the same dimension. G2P2 and GRENADE take a further step by employing contrastive learning strategies. G2P2 enhances the granularity of contrastive learning by jointly training its graph encoder and text encoder with three graph-based contrast strategies (text-node interaction, text-summary interaction, and node-summary interaction) during the pretraining phase to jointly train the graph encoder and text encoder. This allows for the alignment of graph node and text representations in a bimodal embedding space, enabling better capture of fine-grained semantic information in the text while leveraging graph structures to enhance classification model performance. On the other hand, GRENADE jointly optimizes the pre-trained language model encoder and the graph neural network encoder through two self-supervised learning algorithms including graph-centered contrastive learning~\cite{you2021graphcontrastivelearningaugmentations} and graph-centered knowledge alignment.

In order to achieve efficient node classification on textual graphs, GraD~\cite{mavromatis2023traingnnteachergraphaware} employs the concept of knowledge distillation. The core idea is to transfer the graph structure information from the GNN teacher model to the graph-free student model through the distillation process. The student model does not need to use the graph structure during reasoning, thus significantly improving the reasoning efficiency and realizing efficient and accurate node classification. In order to introduce between models with different coupling strengths and flexibility, GraD also proposes three different optimization strategies, namely GraD-Joint, GraD-Alt, and GraD-JKD. Similarly, the THLM~\cite{zou2023pretraininglanguagemodelstextattributed} framework combines BERT with the heterogeneous graph neural network R-HGNN~\cite{Yu_2022}, and through topology-aware pretraining tasks and text augmentation strategies, it pretrains on Text-Attributed Heterogeneous Graphs (TAHGs). After the pretraining phase, the THLM framework retains only the language model for downstream tasks and no longer relies on the auxiliary HGNN, ensuring efficiency and flexibility in processing downstream tasks. Some studies focus on combining LLMs and GNNs in the context of TAGs to reduce training complexity and memory consumption while maintaining the model’s expressiveness. GraphAdapter~\cite{huang2024gnngoodadapterllms} combines GNNs and LLMs specifically for processing TAGs. It firstly uses GNNs for each node in the TAG to model its structural information, then integrates the structural information with context-hidden states of LLMs, and finally transforms the original task into a next-word prediction task by adding task-specific prompts. Its lightweight design, residual connections, and task-related prompts enable the method to exhibit high performance in various downstream tasks, validating its effectiveness in TAG modeling. ENGINE~\cite{zhu2024efficienttuninginferencelarge} through a tunable bypass structure—G-Ladder—combining LLMs and GNNs for efficient fine-tuning and reasoning on TAGs. The framework significantly reduces memory and computational costs while preserving structural information through the introduced lightweight G-Ladder structure that adds tunable parameters next to each layer of LLMs. To further improve efficiency, a caching mechanism is also introduced to precompute the node embeddings during the training process, and dynamic early stop is used to accelerate model inference during the reasoning process.

Compared with GNN-driving-LLM and LLM-driving-GNN, the GNN-LLM-co-driving mode emphasizes the deep interaction and complementarity between the GNNs and LLMs. In this mode, GNN and LLM alternate and enhance each other in the learning process, thereby demonstrating higher robustness and generalization in the integrated processing of graph structure and text information. This co-driving strategy can solve complex tasks efficiently by integrating the graph structure processing capability of GNNs and the text comprehension capability of LLMs.

\begin{table}[tbp]
    \centering
    \caption{A summary of LLMs used in the field of graph mining, highlighting the model architecture, which includes the LLM model, the GNN model, whether the parameters of the LLM are fine-tuned, the predictor of the architecture, the types of datasets, and the downstream task types. In the "Task" column, "Node" denotes node-level tasks, "Link" denotes edge-level tasks, and "Graph" denotes graph-level tasks.}
    \resizebox{\linewidth}{!}{ 
		\begin{tabular}{cccccccc}
        \toprule
         & \multirow{2}{*}{Model} & \multicolumn{4}{c}{Architecture} & \multicolumn{2}{c}{Graph Data}\\
        & & LLMs & GNNs & Finetune & Predictor & Dataset Type & Task\\
        \midrule
        \multirow{4}{*}{\rotatebox{90}{GNN-driving-LLM}} & LLMRec~\cite{wei2024llmreclargelanguagemodels} & ChatGPT & LightGCN & $\times$ & GNN & General & Recommendation \\
        & RLMRec~\cite{Ren_2024} &  ChatGPT,text-embedding-ada-002 & LightGCN & \checkmark & GNN & General & Recommendation\\
        & TAPE~\cite{he2024harnessingexplanationsllmtolminterpreter} &  ChatGPT,Llama2 & RevGAT & $\times$ & GNN & TAGs & Node \\
        & PRODIGY~\cite{huang2023prodigyenablingincontextlearning} &  RoBERTa, MPNet & GCN,GAT & \checkmark & GNN & TAGs & Node \\
        & ALL-in-one~\cite{sun2023onemultitaskpromptinggraph} &  GPT-3.5,DeBERTa & RevGAT  & \checkmark & GNN/LLM & TAGs & Node\\
        & OFA~\cite{liu2024alltraininggraphmodel} &  llama2,e5-large-v2 & R-GCN & $\times$ & GNN & TAGs & Node, Link, Graph\\
        & GaLM~\cite{xie2023graphawarelanguagemodelpretraining} & BERT & RGCN,RGAT & \checkmark & GNN & heterogeneous & Node, Link, edge classification\\
        & LEADING~\cite{xue2023efficientlargelanguagemodels} &  BERT,DeBERTa & GCN,GAT & \checkmark & GNN & TAGs & Node \\
        & AdsGNN~\cite{li2021adsgnnbehaviorgraphaugmentedrelevance} & BERT & GAT & \checkmark & GNN & Heterogeneous & Relevance Prediction, AD recommendation\\
        & TextGNN~\cite{Zhu_2021} & BERT & GAT & \checkmark & GNN & Heterogeneous & Relevance Prediction, AD recommendation \\
        & GraphEdit~\cite{guo2024grapheditlargelanguagemodels} & Vicuna-v1.5 & GCN & \checkmark & Edge Predictor & TAGs & Node, Graph, edge classification\\
        & LLM4NG~\cite{yu2024leveraginglargelanguagemodels} &  ChatGPT & GCN,GAT & $\times$ & GNN & TAGs & Node \\
        & LLM-GNN\cite{chen2024labelfreenodeclassificationgraphs} & gpt-3.5-turbo-0613 & GCN & $\times$ & GNN & TAGs & Node \\
        & OpenGraph~\cite{xia2024opengraphopengraphfoundation} & GPT-4 & Scalable Graph Transformer & $\times$ & Graph Transformer & TAGs, Heterogeneous & Node, Link \\
        
        \midrule
        \multirow{4}{*}{\rotatebox{90}{LLM-driving-GNN}}  & Fatemi et al.~\cite{fatemi2023talklikegraphencoding} &  PaLM/PaLM 2 & - & $\times$ & LLM & Synthetic & Graph\\
        & GPT4Graph~\cite{guo2023gpt4graphlargelanguagemodels} &  GPT-3 & - & $\times$ & LLM & Synthetic, KG & Structural and Semantic Understanding Tasks \\
        & GraphText~\cite{zhao2023graphtextgraphreasoningtext} &  GPT-4 & - & $\times$ & LLM & General,TAGs & Node \\
        & Liu et al.~\cite{liu2023evaluatinglargelanguagemodels} &  GPT, CalderaAI/30B-Lazarus, et al. & - & $\times$ & LLM & Synthetic & Graph Reasoning\\
        & GraphGPT~\cite{tang2024graphgptgraphinstructiontuning} &  Vicuna & Graph Transformer & \checkmark & LLM & TAGs & Node, Link, Graph Match\\
        & MoleculeSTM~\cite{cao2023instructmolmultimodalintegrationbuilding} &  Vicuna-7B & GIN & \checkmark & LLM & TAGs & Graph \\
        & DGTL~\cite{qin2024disentangledrepresentationlearninglarge} &  LLama-2 & GCN, RGAT & \checkmark & LLM & TAGs & Node \\
        & graphtranslator~\cite{zhang2024graphtranslatoraligninggraphmodel} & ChatGLM2-6B & GraphSAGE & $\times$ & LLM & TAGs & Node, KG Question Answering\\
        & HiGPT~\cite{tang2024higptheterogeneousgraphlanguage} & GPT-3.5 & HetGNN, HAN, HGT & $\times$ & LLM & Heterogeneous & Node, Graph, Relation Prediction \\
        & GIMLET~\cite{zhao2023gimletunifiedgraphtextmodel} &  T5 & - & \checkmark & LLM & Synthetic & Molecular Property Prediction\\
        & MuseGraph~\cite{tan2024musegraphgraphorientedinstructiontuning} &  GPT-4 & - & \checkmark & LLM & General & Node, Link \\
        & InstructGLM~\cite{ye2024languagegraphneeds} &  T5,Llama-7b & - & \checkmark & LLM & TAGs & Node, Link \\
        & GraphLLM~\cite{chai2023graphllmboostinggraphreasoning} &  LLaMA 2 & Graph Transformer & \checkmark & LLM & KG & Graph Reasoning \\
        & Graph-ToolFormer~\cite{zhang2023graphtoolformerempowerllmsgraph} &  GPT-J & Graph-Bert,SEG-Bert & \checkmark & LLM & General,Synthetic,TAGs & Graph Reasoning \\
        & GPT4GNAS~\cite{wang2024graphneuralarchitecturesearch} &  GPT-4 & GCN,GAT et al. & $\times$ & LLM & General &  Graph Neural Architecture Search \\
        & ChatRule~\cite{luo2024chatrulemininglogicalrules} & ChatGPT & - & $\times$ & LLM & KG & KG Reasoning\\
        & GNP~\cite{tian2023graphneuralpromptinglarge} & FLAN-T5 & GAT & \checkmark & LLM & KG & Commonsense and Biomedical Reasoning\\
        
        \midrule
        \multirow{4}{*}{\rotatebox{90}{GNN-LLM-co-driving}} 
        & GraphFormers~\cite{yang2023graphformersgnnnestedtransformersrepresentation} & UniLM-base & GNN Components & \checkmark & GNN, LLM & TAGs & Link \\
        & PATTON~\cite{jin2023pattonlanguagemodelpretraining} & BERT, SciBERT & GraphFormers & \checkmark & LLM & Heterogeneous & Node, Retrieval, Re-ranking for Link Prediction\\
        & GLEM~\cite{zhao2023learninglargescaletextattributedgraphs} & DeBERTa & RevGAT & \checkmark & GNN, LLM & TAGs & Node \\
        & GREASELM~\cite{zhang2022greaselmgraphreasoningenhanced} & RoBERTa-Large, AristoRoBERTa, SapBERT & GAT & \checkmark & GNN, LLM & KG & Multiple Choice Question Answering \\
        & Text2Mol~\cite{edwards-etal-2021-text2mol} & SciBERT & GCN & \checkmark & GNN, LLM & General  & Cross-modal Information Retrieval\\
        & MoleculeSTM~\cite{liu2024multimodalmoleculestructuretextmodel} & SciBERT & GIN & \checkmark & GNN, LLM & TAGs & Retrieval, Molecular Editing\\
        & CLAMP~\cite{seidl2023enhancingactivitypredictionmodels} & BioBERT, Sentence-T5, KV-PLM, etc. & GCN, GIN & \checkmark & GNN, LLM & TAGs & Bioactivity Prediction \\
        & ConGraT~\cite{brannon2024congratselfsupervisedcontrastivepretraining} & DistilGPT2, all-mpnet-base-v2 & GAT & \checkmark & GNN, LLM & General & Representation Learning\\
        & G2P2~\cite{wen2024prompttuninggraphaugmentedlowresource} & BERT & GCN & \checkmark & GNN, LLM & General & Text Classification\\
        & GRENADE~\cite{li2023grenadegraphcentriclanguagemodel} & BERT & SAGE, RevGAT-KD, etc. & \checkmark & GNN, PLM & TAGs & Node, Link \\
        & GraD~\cite{mavromatis2023traingnnteachergraphaware} & BERT, SciBERT & GraphSAGE & \checkmark & LLM & TAGs, Heterogeneous & Node \\
        & THLM~\cite{zou2023pretraininglanguagemodelstextattributed} & BERT & R-HGNN & \checkmark & LLM & TAHGs & Node, Link \\
        & GraphAdapter~\cite{huang2024gnngoodadapterllms} & Llama 2, RoBERTa, GPT-2 & GraphSAGE & $\times$ & GNN, LLM & TAGs & Node \\
        & ENGINE~\cite{zhu2024efficienttuninginferencelarge} & LLaMA2-7B, e5-large & GCN,SAGE,GAT & $\times$ & GNN, LLM & TAGs & Node \\
        \bottomrule
    \end{tabular}}
    \label{tab:models_overview}
\end{table}

\section{Summary and Discussion Analysis}

\subsection{Summary}

The previous sections have discussed the related work on graph mining domain modeling using the Large Language Models. Through a systematic review, it is evident that the combination of LLMs and GNNs brings new research ideas and directions for graph mining. Specifically, LLM demonstrates strong capabilities in semantic understanding and text feature extraction, while GNN is uniquely suited to capture graph structure and complex node relationships. Therefore, various studies have integrated the two models through multiple architectures and strategies to enhance the performance of graph mining tasks. The models listed in Table \ref{tab:models_overview} are categorized and compared according to their main driving approaches, summarizing their performance across tasks, datasets, performance metrics, and applicable scenarios.

\begin{figure}[tbp]
\centering
\includegraphics[width=10.5 cm]{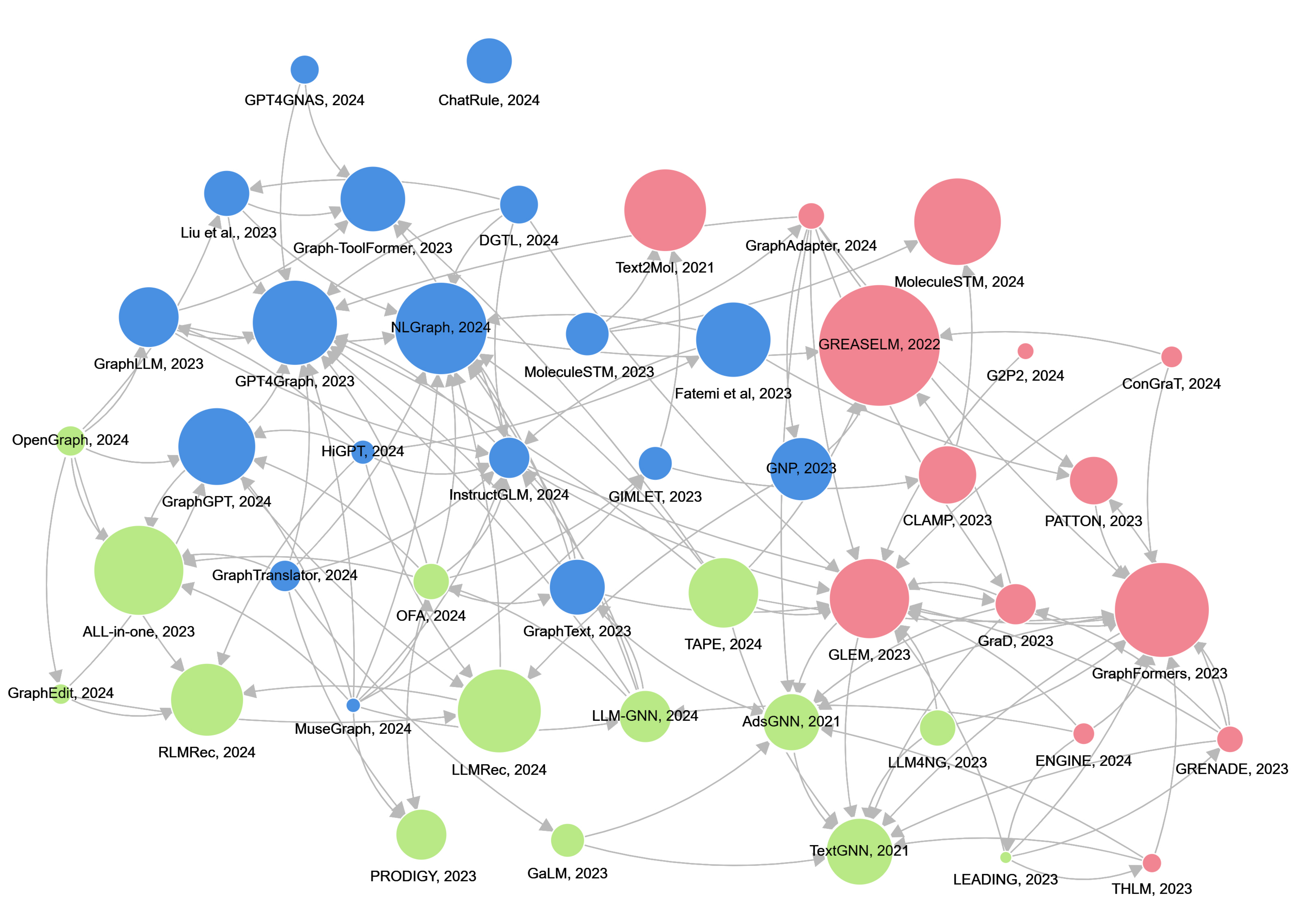}
\caption{The citation network uses directed edges to represent the citation relationships between articles. Green, blue, and red nodes represent literature categorized as “GNN-driving-LLM”, “LLM-driving-GNN” and “GNN-LLM-co-driving” respectively. The size of the nodes reflects the number of citations of each paper. }
\label{fig:graph}
\end{figure}
\unskip

In the meantime, to facilitate readers to understand the relationships between the literature, we innovatively sketch a citation network graph of the important literature discussed in this review, as shown in figure \ref{fig:graph}, using directed edges to represent the citation relationship between articles. 
In the figure, green, blue, and red nodes represent the literature categorized as “GNN-driving-LLM”, “LLM-driving-GNN” and “GNN-LLM-co-driving”, respectively; the size of the nodes reflects the number of citations.  
It can be observed that there are certain citation relationships among different categories, which indicates that the research in these three categories has strong intersectionality and complementarity in terms of methodology and application scenarios. It is worth noting that the citation relationship of the green node (GNN-driving-LLM) exhibits a relatively scattered citation relationship, which indicates that the researches in this direction emphasize the independent application in specific scenarios, with less interaction with other categories. Each category has at least one representative paper (with the highest number of citations).
For instance, the representative work of “LLM-driving-GNN” is NLGraph~\cite{wang2024languagemodelssolvegraph}, and the representative work of “GNN-LLM-co-driving” is GREASELM~\cite{zhang2022greaselmgraphreasoningenhanced}. Overall, the graph intuitively visualizes the synergistic and evolutionary relationships between different research directions in the field, providing a valuable reference for future research.

\subsection{Discussion Analysis}

\begin{table}[tbp]
    \centering
    \caption{Summary of experimental setups on selected models.\label{tab:evaluate}}
    \resizebox{\linewidth}{!}{ 
		\begin{tabular}{cccccccc}
        \toprule
        & Models & LLMs & GNNs & Environment & Datasets & ratio* & Code\\
        \midrule
        \multirow{4}{*}{\rotatebox{90}{Node Classification}} & TAPE-GNN-$h_{TAPE}$~\cite{he2024harnessingexplanationsllmtolminterpreter} &  DeBERTa-base & RevGAT & 4$\times$ Nvidia RTX A5000 24GB GPUs & ogbn-arxiv & 77.50 ± 0.12 & \href{https://github.com/XiaoxinHe/TAPE}{Link}  \\
        & GLEM-G~\cite{zhao2023learninglargescaletextattributedgraphs} &  DeBERTa-base & RevGAT & 4$\times$ Nvidia RTX A5000 24GB GPUs & ogbn-arxiv & 76.57 ± 0.29 & \href{https://github.com/AndyJZhao/GLEM}{Link} \\
        & OFA~\cite{liu2024alltraininggraphmodel} &  llama2-13b & R-GCN & - & ogbn-arxiv & 77.51±0.17 & \href{https://github.com/LechengKong/OneForAll}{Link}\\
        & InstructGLM~\cite{ye2024languagegraphneeds} &  Llama-7b & - & 4$\times$ 40G Nvidia A100 GPUs & ogbn-arxiv & 75.70 ± 0.12 & \href{https://github.com/Graphlet-AI/llm-graph-ai}{Link}\\
        & GraphGPT-7B-v1.5-stage2~\cite{tang2024graphgptgraphinstructiontuning} &  Vicuna & Graph Transformer & 4$\times$ 40G Nvidia A100 GPUs & ogbn-arxiv & 75.11 & \href{https://github.com/HKUDS/GraphGPT}{Link}\\
        & GRENADE~\cite{li2023grenadegraphcentriclanguagemodel} & BERT & RevGAT-KD & - & ogbn-arxiv & 76.21±0.17 & \href{https://github.com/bigheiniu/GRENADE}{Link}\\
        & GraDBERT~\cite{mavromatis2023traingnnteachergraphaware} & BERT & GraphSAGE & - & ogbn-arxiv & 75.05±0.11 & \href{https://github.com/cmavro/GRAD}{Link}\\
        & GraphAdapter~\cite{huang2024gnngoodadapterllms} & Llama 2 & GraphSAGE & Nvidia A800 80GB GPU & Ogbn-arxiv & 77.07±0.15 & \href{https://github.com/hxttkl/GraphAdapter}{Link}\\
        & ENGINE~\cite{zhu2024efficienttuninginferencelarge} & LLaMA2-7B & GCN,SAGE,GAT & 6$\times$ Nvidia RTX 3090 GPUs & Ogbn-arxiv & 76.02±0.29 & \href{https://github.com/ZhuYun97/ENGINE}{Link} \\
        & PATTON~\cite{jin2023pattonlanguagemodelpretraining} & BERT, SciBERT & GraphFormers & 4$\times$ Nvidia A6000 GPUs & Amazon-Sports & 78.60±0.15 & \href{https://github.com/PeterGriffinJin/Patton}{Link}\\
        & THLM~\cite{zou2023pretraininglanguagemodelstextattributed} & BERT & R-HGNN & 4$\times$ Nvidia RTX 3090 GPUs & GoodReads & 81.57 & \href{https://github.com/Hope-Rita/THLM}{Link} \\
        
        \midrule
        \multirow{4}{*}{\rotatebox{90}{**}} 
        & LLMRec~\cite{wei2024llmreclargelanguagemodels} & gpt-3.5-turbo, text-embedding-ada-002 & LightGCN & Nvidia RTX 3090 GPU & MovieLens & 36.43 & \href{https://github.com/HKUDS/LLMRec}{Link} \\
        & RLMRec~\cite{Ren_2024} &  gpt-3.5-turbo, text-embedding-ada-002 & LightGCN & Nvidia RTX 3090 GPU & Amazon-book & 14.83 & \href{https://github.com/HKUDS/RLMRec}{Link}\\
        \bottomrule
		\end{tabular}
        }
	\noindent{\footnotesize{*In the ratio, node classification tasks are evaluated by accuracy, while recommendation tasks are evaluated by recall. **Recommendation.}}
\end{table}

When evaluating models that combine LLMs with GNNs, the choice of test environment and test data is crucial to ensure the reliability and fairness of the results. 
In this section, the test environments of existing models and the selection of test data are summarized and described, and their comparison is shown in Table \ref{tab:evaluate}.  
As different test datasets are used in various studies covering multiple graph mining tasks, such as node classification, link prediction, and graph classification, in order to facilitate the comparison, we try to select a unified dataset with downstream task types as the classification. Due to the variety of graph-level tasks, it is inconvenient to perform comparative analysis, and only the data of node classification and recommendation tasks are selected for presentation here. In addition, to ensure the reproducibility of the results, we only selected the open source data.
The table lists the hardware configurations (e.g., GPU model and number) used for each model,  the specific test datasets, as well as the best-performing large language model and graph neural network model in the test. The organization of this information can help readers understand each model's computational efficiency and applicability and also provides a reliable reference for subsequent research for performance comparison and technology validation under different experimental conditions.

During the collation process, we observed that all models show improvements relative to the baseline, which powerfully demonstrates that combining LLMs and GNNs holds more incredible promise than using either model individually. Specifically, combining the robust language understanding and generation capabilities of LLMs with the graph-structured data processing strengths of GNNs results in a significant improvement in overall performance. This cross-model synergy not only improves the understanding of complex graph data but also enhances the model's performance in tasks such as node classification, showing great potential for joint applications. This finding provides important insights for future research, emphasizing the importance of multi-model collaboration.

However, while we observe performance improvements across models for specific tasks, the overall improvement remains modest. For example, the recently proposed TAPE~\cite{he2024harnessingexplanationsllmtolminterpreter} model only improves by $1.6\%$ over the baseline. This phenomenon may stem from the fact that the current method of combining LLMs and GNNs is too simple, which is more of a "stitching" rather than a true deep fusion. 
This simple combination fails to fully utilize the respective strengths of the two models, nor does it achieve true complementarity between them in terms of feature extraction and representation learning. As a result, despite some degree of performance improvement, the expected significant optimization is not achieved. This suggests that future research must explore more complex and efficient integration strategies to achieve deep synergy between LLMs and GNNs, thereby driving further improvement in model performance.

\section{Future Direction}

Based on the above analysis, it is clear that there are still many directions in this research area that have yet to be fully explored and deeply understood. Therefore, this section will further analyze these issues, focusing on the drawbacks and potential research opportunities in the current study, with a view to providing new ideas and insights for future academic exploration.

\subsection{Multimodal Graph Data Processing}

In graph data, nodes may be enriched with information in multiple modalities, such as text, images, and videos. These modalities may contain rich information, so understanding these multimodal data can help improve graph learning. A number of recent studies have explored the ability of LLMs to process and integrate multimodal data, and these studies have shown that LLMs exhibit significant capabilities in this area ~\cite{wu2024nextgptanytoanymultimodalllm, liu2023visualinstructiontuning}, which makes it possible to apply LLMs to multimodal graph data. Future research will focus on exploring how to design a unified model to jointly encode data in different modalities such as graphs, text, and images. This will be applied to areas such as social network analysis, product recommendation, and molecular modeling to enhance the performance of models in complex, multimodal scenarios.

\subsection{Addressing the Hallucination Problem in Large Language Models}

While LLMs have shown a fantastic ability to generate text, they are prone to hallucinations and misinformation due to the fact that they tend to generate answers in a single pass and lack the ability to adjust dynamically. Hallucination means that the information generated by the model seems reasonable but is actually inaccurate, deviating from the user input, context, or even from the facts~\cite{zhang2023sirenssongaiocean}. In specific sophisticated fields, such misinformation is unacceptable. Therefore, future research directions should focus on solving the hallucination problem and reducing the generation of misinformation, and this can be accomplished with the help of graph data. For example, it can be done by combining external knowledge graphs so that the big language model can reason step-by-step in generating answers and refer to reliable structured data sources to verify the accuracy of the information. Furthermore, using multi-hop reasoning and dynamic knowledge retrieval mechanisms enables the model to continuously adjust and correct its output according to the context, thus providing more accurate and trustworthy answers. By employing these strategies, the model will be more stable and reliable, especially in application scenarios that require high accuracy.

\subsection{Enhancing the Capability to Solve Complex Graph Tasks}

Currently, LLMs are primarily applied to basic graph tasks such as node classification and link prediction, but the remarkable capabilities that LLMs demonstrate in various areas suggest that their potential in graph data extends beyond these tasks. As a result, more and more research has explored their application to more complex graph tasks, such as graph generation ~\cite{yao2024exploringpotentiallargelanguage}, question answering over knowledge graph ~\cite{huang2019knowledge} and knowledge graph construction ~\cite{ peng2023knowledgegraphsopportunitieschallenges}. LLMs can be used to generate novel molecular structures, analyze complex relationship patterns in social networks, or assist in constructing more contextually connected knowledge graphs. The solution to these complex tasks will drive the further development of LLMs in a variety of fields, including biomedicine, social network analysis, and natural language processing.

\section{Conclusion}

In recent years, significant progress has been made in the application of LLMs in the field of graph mining. This study aims to provide an overview, summarize the research in this area, and provide potential directions for future research. We propose a new taxonomy based on different driving modes: GNN-driving-LLM, LLM-driving-GNN and GNN-LLM-co-driving. Each mode exhibits unique advantages and application potentials, especially when dealing with complex graph structures and textual information. The combination of LLMs and GNNs has brought new opportunities to graph mining. The semantic understanding capability of LLMs complements the structural information processing capability of GNNs, significantly improving the effectiveness of graph mining tasks. Despite the many opportunities presented by the combination of GNNs and LLMs, their high computational demands and model complexity remain challenges. Future research should explore optimizing the integration model of GNNs and LLMs to achieve more powerful graph learning and reasoning capabilities while ensuring computational efficiency, thus advancing the field of graph mining.
\bibliographystyle{unsrt}  
\bibliography{references}

\end{document}

%% file: fig/tree.tex
\tikzstyle{root}=[draw=black,
    rounded corners = 3pt, minimum height=1em,
    fill=output-white!40, text opacity=1, align=center,
    fill opacity=.5,  text=black, align=left, font=\scriptsize,
    inner xsep=3pt,
    inner ysep=1pt,
]

\tikzstyle{middle}=[draw=black,
    rounded corners = 3pt, minimum height=1em,
    fill= CornflowerBlue!20, text opacity=1, align=center,
    fill opacity=.5,  text=black, align=left, font=\scriptsize,
    inner xsep=3pt,
    inner ysep=1pt,
]

\tikzstyle{leaf}=[draw=black,
    rounded corners = 3pt, minimum height=1em,
    fill= CornflowerBlue!40, text opacity=1, align=center,
    fill opacity=.5,  text=black, align=left, font=\scriptsize,
    inner xsep=3pt,
    inner ysep=1pt,
]

\begin{figure*}[tbp]
\centering
\begin{forest}
  for tree={
  forked edges,
  grow=east,
  reversed=true,
  anchor=base west,
  parent anchor=east,
  child anchor=west,
  base=middle,
  font=\scriptsize,
  rectangle,
  line width=0.7pt,
  draw=output-black,
  rounded corners = 3pt, align=left,
  minimum width=2em,
    s sep=6pt,
    inner xsep=3pt,
    inner ysep=1pt,
  },
  where level=1{text width=4.5em}{},
  where level=2{text width=6em,font=\scriptsize}{},
  where level=3{font=\scriptsize}{},
  where level=4{font=\scriptsize}{},
  where level=5{font=\scriptsize}{},
  [Large Language Models Meets Graph Mining, root, rotate=90, anchor=north, edge=output-black
    [GNN-driving-LLM, middle, edge=output-black, text width=7em
        [LLMs for Feature Enhancement, middle, text width=10.8em, edge=output-black
            [LLMRec~\cite{wei2024llmreclargelanguagemodels}{,} RLMRec~\cite{Ren_2024}{,} TAPE~\cite{he2024harnessingexplanationsllmtolminterpreter}{,} PRODIGY~\cite{huang2023prodigyenablingincontextlearning}{,} \\ALL-in-one~\cite{sun2023onemultitaskpromptinggraph}, leaf, text width=21em, edge=output-black]
        ]
        [LLMs for Text Embedding Output, middle, text width=10.8em, edge=output-black
            [OFA~\cite{liu2024alltraininggraphmodel}{,} GaLM~\cite{xie2023graphawarelanguagemodelpretraining}{,} LEADING~\cite{xue2023efficientlargelanguagemodels}{,} AdsGNN~\cite{li2021adsgnnbehaviorgraphaugmentedrelevance}{,} TextGNN~\cite{Zhu_2021}{,} \\GraphEdit~\cite{guo2024grapheditlargelanguagemodels}, leaf, text width=21em, edge=output-black]
        ]
        [LLMs for Data Augmentation, middle, text width=10.8em, edge=output-black
            [LLM4NG~\cite{yu2024leveraginglargelanguagemodels}{,} LLM-GNN\cite{chen2024labelfreenodeclassificationgraphs}{,} OpenGraph~\cite{xia2024opengraphopengraphfoundation}, leaf, text width=21em, edge=output-black]
        ]
    ]
    [LLM-driving-GNN, middle, edge=output-black, text width=7em
        [Graph Encoded as Sequences, middle, text width=10.8em, edge=output-black
            [GraphText~\cite{zhao2023graphtextgraphreasoningtext}{,} Fatemi et al.~\cite{fatemi2023talklikegraphencoding}{,} GPT4Graph~\cite{guo2023gpt4graphlargelanguagemodels}{,} DGTL~\cite{qin2024disentangledrepresentationlearninglarge}{,} \\NLGraph\cite{wang2024languagemodelssolvegraph}{,} Liu et al.~\cite{liu2023evaluatinglargelanguagemodels}{,} GraphGPT~\cite{tang2024graphgptgraphinstructiontuning}{,} MoleculeSTM~\cite{cao2023instructmolmultimodalintegrationbuilding}{,}\\ GraphTranslator~\cite{zhang2024graphtranslatoraligninggraphmodel}{,} HiGPT~\cite{tang2024higptheterogeneousgraphlanguage}, leaf, text width=21em, edge=output-black]
        ]
        [Fine-tuned LLMs, middle, text width=10.8em, edge=output-black
            [GIMLET~\cite{zhao2023gimletunifiedgraphtextmodel}{,} MuseGraph~\cite{tan2024musegraphgraphorientedinstructiontuning}{,} InstructGLM~\cite{ye2024languagegraphneeds}{,} GraphLLM~\cite{chai2023graphllmboostinggraphreasoning}, leaf, text width=21em, edge=output-black]
        ]
        [Other Scenarios, middle, text width=10.8em, edge=output-black
            [Graph-ToolFormer~\cite{zhang2023graphtoolformerempowerllmsgraph}{,} GPT4GNAS~\cite{wang2024graphneuralarchitecturesearch}{,} ChatRule~\cite{luo2024chatrulemininglogicalrules}{,} GNP~\cite{tian2023graphneuralpromptinglarge}, leaf, text width=21em, edge=output-black]
        ]
    ]
    [GNN-LLM-co-driving, middle, edge=output-black, text width=7em
        [Fusion of GNNs and LLMs, middle, text width=10.8em, edge=output-black
            [GraphFormers~\cite{yang2023graphformersgnnnestedtransformersrepresentation}{,} PATTON~\cite{jin2023pattonlanguagemodelpretraining}{,} GLEM~\cite{zhao2023learninglargescaletextattributedgraphs}{,} GREASELM~\cite{zhang2022greaselmgraphreasoningenhanced}, leaf, text width=21em, edge=output-black]
        ]
        [Encoding and Alignment, middle, text width=10.8em, edge=output-black
            [Text2Mol~\cite{edwards-etal-2021-text2mol}{,}  MoleculeSTM~\cite{liu2024multimodalmoleculestructuretextmodel}{,} CLAMP~\cite{seidl2023enhancingactivitypredictionmodels}{,} ConGraT~\cite{brannon2024congratselfsupervisedcontrastivepretraining}{,} \\G2P2~\cite{wen2024prompttuninggraphaugmentedlowresource}{,} GRENADE~\cite{li2023grenadegraphcentriclanguagemodel}, leaf, text width=21em, edge=output-black]
        ]
        [Other Scenarios, middle, text width=10.8em, edge=output-black
            [GraD~\cite{mavromatis2023traingnnteachergraphaware}{,} THLM~\cite{zou2023pretraininglanguagemodelstextattributed}{,} GraphAdapter~\cite{huang2024gnngoodadapterllms}{,} ENGINE~\cite{zhu2024efficienttuninginferencelarge}, leaf, text width=21em, edge=output-black]
        ]
    ]
  ]
\end{forest}
\caption{The proposed taxonomy on LLM-GNN-combined techniques.}
\label{fig:tree}
\end{figure*}